\documentclass[10pt]{article} 
\usepackage[accepted]{tmlr}

\usepackage{amsmath,amsfonts,bm}

\def\eqref#1{equation~\ref{#1}}

\def\1{\bm{1}}

\DeclareMathAlphabet{\mathsfit}{\encodingdefault}{\sfdefault}{m}{sl}
\SetMathAlphabet{\mathsfit}{bold}{\encodingdefault}{\sfdefault}{bx}{n}

\DeclareMathOperator*{\argmax}{arg\,max}

\usepackage{hyperref}
\usepackage{url}
\usepackage{microtype}
\usepackage{graphicx}
\usepackage{booktabs} 
\usepackage{placeins}
\usepackage{subcaption}
\usepackage{amsmath}
\usepackage{amssymb}
\usepackage{mathtools}
\usepackage{amsthm}
\usepackage{todonotes}

\title{Cyclophobic Reinforcement Learning}

\author{\name Stefan Wagner$^{1}$ \  \name Peter Arndt$^{1}$ \ \name Jan Robine$^{2}$ \ \name Stefan Harmeling$^{2}$ \\
  \addr $^{1}$Heinrich Heine University Düsseldorf \ \addr $^{2}$Technical University Dortmund \\
  $\{$stefan.wagner, peter.arndt$\}$@hhu.de \  $\{$jan.robine, stefan.harmeling$\}$@tu-dortmund.de}

\def\month{08}  
\def\year{2023} 
\def\openreview{\url{https://openreview.net/forum?id=83}} 

\begin{document}

\maketitle

\begin{abstract}
  In environments with sparse rewards, finding a good inductive bias for
  exploration is crucial to the agent's success. However, there are two
  competing goals: novelty search and systematic exploration. While existing
  approaches such as curiosity-driven exploration find novelty, they sometimes
  do not systematically explore the whole state space, akin to
  depth-first-search vs breadth-first-search. In this paper, we propose a new
  intrinsic reward that is cyclophobic, i.e., it does not reward novelty, but
  punishes redundancy by avoiding cycles. Augmenting the cyclophobic intrinsic
  reward with a sequence of hierarchical representations based on the agent's
  cropped observations we are able to achieve excellent results in the MiniGrid
  and MiniHack environments. Both are particularly hard, as they require complex
  interactions with different objects in order to be solved. Detailed
  comparisons with previous approaches and thorough ablation studies show that
  our newly proposed cyclophobic reinforcement learning is more sample efficient
  than other state of the art methods in a variety of tasks. 
\end{abstract}

\section{Introduction}
Exploration is one of reinforcement learning's most important problems. Learning
success largely depends on whether an agent is able to explore its environment
efficiently. Random exploration explores all possibilities but at great costs,
since it possibly revisits states very often. More efficient approaches can use
intrinsic rewards based on novelty \citep{pseudo, pseudo_ovr}, where the occurrence of states is counted
and some novelty metric is determined or can be based on curiosity where novelty
in the representations is determined \citep{icm, rnd, ride, noveld}. This often leads to great results,
but at the price of possibly not exploring all corners of the environment
systematically. The latter also paradoxically end up being less sample
efficient, when aleatoric uncertainty in the observations is not large enough to
give an informative reward signal, such as in environments which are challenging
from a combinatorial standpoint, but where the observations are simple
\citep{ioca}. Especially in environments where the task is complex we would
ideally pursue both goals: novelty search and systematical exploration.

How can we favor novelty while ensuring that the environment is systematically
explored? To achieve this, we propose \emph{cyclophobic reinforcement learning}
which is based on the simple idea of avoiding cycles during exploration. More
precisely, we define a negative intrinsic reward, i.e., the cyclophobic
intrinsic reward that penalizes redundancy in the exploration history. With this
we ensure two things: first, the agent receives immediate feedback about which
state-action pair to discourage. Second, by maximizing the agent's actions it is
forced to take the next unexplored state-action pair since already explored
state-action pairs will have a negative intrinsic reward. This approach is
similar to optimistic initialization \citep{dmoptim} where the value function is
initialized with a high value and as the update rule is applied, the values will
decrease for subsequent visits, thus forcing the agent to take the next
unexplored action with highest remaining value. In this sense, the cyclopohobic
intrinsic reward can be seen as a count-based extension of optimistic
initialization.

The cyclophobic intrinsic reward is further enhanced by applying it to several
cropped observations of the environment which we call hierarchical state
representations. Furthermore, we call the single cropped views, hierarchical
views. The notion of redundancy can be defined relative to cropped views of the
agent: while cycles in the global view induce cycles in a corresponding smaller
view, the converse is not the case. E.g., a MiniGrid agent turning four times to
the left produces a cycle in state space that we would like to avoid everywhere.
This cycle occurrs in the global view, but penalizing it only in the global view
does not avoid it in other locations. However, with a hierarchy of views, we
record a cycle in some smaller view as well, which allows us to transfer this
knowledge to any other location of the environment and to hereby avoid further
cycles for parts in the environment that look the same. Similarly, cycles on a
hierarchical level reveal structural properties about the environment. An
interesting object such as a key, produces less cycles in a smaller view than
some other object (since e.g. the key can be interacted with). Thus the
probability of picking up the key increases, since other \emph{less interesting}
observations produce more cycles (e.g a wall). Overall, a hierarchy of cropped
views is a state representation that is complementary to the cyclophobic
intrinsic reward that does not require the learning of an inverse dynamics model
as in other works \citep{icm, e3b}. The approach of exploiting a prioproceptive
inductive bias for the observations can also be seen in \cite{ioca}, where a
360° panoramic view of the environment is created, which enhances the information
the intrinisic reward gives about the environment. An important aspect of
exploration and another desirable property of hierarchical views is that they
are task-agnostic \citep{ioca}. That is, the cropped views of the environment
can be transferred to other environments and are not task dependent. Thus,
hierarchical state representations on one hand generalize cycles when learning
about a task, but also represent task-agnostic knowledge when transferring
knowledge to other tasks.

\textbf{Contributions:}

1. We introduce cyclophobic reinforcement learning for efficient exploration in
environments with extremely sparse rewards and large action spaces. It is based
on a cyclophobic intrinsic reward for systematic exploration applied to a
hierarchy of views. Instead of rewarding novelty, we avoid redundancy by
penalizing cycles, i.e., repeated state-action pairs in the exploration history.
Our approach can be applied to any MDP for which a hierarchy of views can be
defined.
 
2. In the sparse-reward setting of the MiniGrid and MiniHack environments we
thoroughly evaluate cyclophobic reinforcement learning and can show that it
achieves excellent results compared to existing methods, both for tabula-rasa
and transfer learning.
  
3. In an ablation study we provide deeper insights into the interplay of the
cyclophobic intrinsic reward and the hierarchical state representations. We
provide insight into the role of learning representations for exploration. We
show that exploration and therefore learning success depend greatly on the
function approximator learning appropriate invariances that reduce the
complexity of the state space.

\paragraph*{Notation:}
We define an MDP as a tuple
$(\mathcal{S}, \mathcal{A}, \mathcal{P}, \mathcal{R}, \gamma)$, where the agent
and the environment interact continuously at discrete time steps
$t=0,1,2,3, \ldots$ We define the state an agent receives from the environment
as a random variable $S_t \in \mathcal{S}$, where $S_t = s$ is some
representation of the state from the set of states $\mathcal{S}$ at timestep
$t$. From that state, we define a random variable for the agent selecting an
action $A_t \in \mathcal{A}$ where $A_t = a$ is some action in the possible set
of actions $\mathcal{A}$ for the agent at timestep $t$. This action is selected
according to a policy $\pi(a \mid s)$ or $\pi(s)$ if the policy is
deterministic. One time step later as a consequence of its action, the agent
receives a numerical reward which is a random variable $R_{t+1} \in \mathbb{R}$,
where $R_{t+1} = r$ is some numerical reward at timestep $t+1$. Finally, the
agent finds itself in a new state $S_{t+1}$. Furthermore we define a POMDP
$(\mathcal{S}, \mathcal{A}, \mathcal{O}, \mathcal{P}, \mathcal{R}, \mathcal{Z},
\gamma)$ as a generalization of an MDP in the case the true state space
$\mathcal{S}$ is unknown. That is, the agent sees the state $s \in \mathcal{S}$
through an observation $o \in \mathcal{O}$, where an observation function
$\mathcal{Z}: \mathcal{S} \rightarrow \mathcal{O}$ maps the true state to the
agent's
observation. 


\section{Building Blocks}
We begin by first defining the cyclophobic intrinsic reward and hierarchical
state representations as they are the building blocks of our method. Finally, we
define a policy which combines the intrinsic and extrinsic rewards together with
the hierarchical state representations to form a global policy which the agent
acts upon.

\subsection{Cyclophobic Intrinsic Reward}
\label{sec:cir}
For efficient exploration, redundancy must be avoided. A sign for redundancy is
when states are repeatedly explored, with other words, when the agent encounters
cycles in the state space instead of focusing on novel areas. To guide the
exploration, we will penalize cycles using a cycle penalty, which we call
cyclophobic intrinsic reward (a negative intrinsic reward). This avoids
redundancy such that uninteresting parts of the state-action space are discarded
quickly. For instance, if an agents gets stuck in some area, typically, it is
facing numerous cycles. In such a situation we would like the agent to assign
penalties to the repeating state-action pairs in order to focus on more
promising parts of the state space that do not cause immediate repetition.

Formally, let us assume that we have per episode a history of previous
state-action pairs
$\mathcal{H}_\text{episodic}=\{(s_1, a_1), (s_2, a_2), \ldots, (s_t, a_t)\}$ and
we are currently at the state-action pair $(s_{t+1}, a_{t+1})$. We say that we
have encountered a cycle if the state in the current state-action pair appeared
already in the history, i.e.
$(s_{t+1}, a_{t+1}) \in \mathcal{H}_\text{episodic}$. For its first repeated
occurrence, we will penalize the state-action pair $(s_{t}, a_{t})$ (just before
the cycle) by negative one,
\begin{align}
	r_\text{cycle}(s, a;  \mathcal{H}_\text{episodic}) = -1.
\end{align}
Note that the cycle penalty as a function depends on the episodic history
$\mathcal{H}_{\text{episodic}}$. If a cycle is encountered multiple times, e.g.
$l$ times, the overall cycle penalty for a state-action pair is $-l$. That is,
during exploration the cycle penalty can penalize indefinitely. However, note that every
single repeated encounter is penalized with $-1$. For pairs $(s,a)$ that have
not created a cycle, $r_\text{cycle}(s, a; \mathcal{H}_\text{episodic})= 0$.

\paragraph{Learning with cycle penalties.}

In principle, the cycle penalty can be combined with any reinforcement learning
algorithm. However, we explain how it can be built into the SARSA update rule,
since the latter will propagate the penalty across the trajectory. This is
important as we want the cycle discouragement to happen on-policy, i.e., the
state-action pair before the cycle should be penalized. Eventually, through the
SARSA updates this information will propagate to other state-action
pairs. 
To learn the Q-function, we are employing the standard SARSA update rule,
\begin{align}
	Q(s, a) &\leftarrow (1-\eta) \ Q(s, a) + \eta \big[r(s,a) + \gamma\, Q(s', a')\big]
            \label{eq:qfunction}
\end{align}
with $(s, a, r_\text{ex}, s', a')$ being a transition, $\eta$ being step size
and where the total reward $r(s,a)$ for the state-action pair $(s,a)$ is the
weighted sum of the extrinsic reward from the environment $r_\text{ex}$ and the
cyclophobic intrinsic reward $r_\text{cycle}$
defined above,
\begin{align}
	r(s,a) =  \rho\, r_\text{ex}(s,a) + r_\text{cycle}(s,a;  \mathcal{H}_\text{episodic}).
\end{align}
where $\rho$ trades off extrinsic and intrinsic rewards.

\subsection{Hierarchical State Representations}
\label{sec:hsr}
Besides penalizing cycles, the second key idea of this paper is to consider a
hierarchy of state representations. For this, we repeatedly crop the agent’s
observations to induce additional partially observable Markov decision processes
(POMDPs). In general, restricting the view leads to ignoring information about
the environment, however in combination with the cyclophobic intrinsic reward we
gain additional information about the structure of the environment as we get
different Q-functions for different POMDP's. The relevant insight is that on
limited views lower down the hierarchy, trajectories can contain cycles that
have not been experienced on views higher up the hierarchy. These cycles in
smaller views represent transferable knowledge about the structure of the
environment. E.g. in the MiniGrid environment (see Section
\ref{sec:minigrid-experiments}) encountering a wall in the smallest view will
cause a cycle, capturing that running into a wall is counterproductive. On
larger views this knowledge is only available directly, if we tried all walls
everywhere. So, the smaller views capture relevant invariances that also apply
to the larger views.
Hierarchical state representations also enable task-agnostic learning
\citep{ioca}. That is, when learning an exploration policy we want to learn a
policy that together with its representations can be transferred to other
environments. By definition, the hierarchical state representations allow us to
transfer learned invariances in smaller views to other environments by simply
keeping these views for transfer learning. In general MDPs, it might not be
obvious how to define hierarchical state representations. Ideally, the agent has
some sort of sense of ``location'' to allow cropped views to be meaningful. In
general, for grid world-like environments, the hierarchical state
representations are easily definable, since the agent has a defined location
that can be used to define smaller neighborhoods typically corresponding to
limited views around the agent or from the agent's view.
\begin{figure*}[htb!]
	\centering
  \begin{subfigure}{.35\textwidth}		\centering \includegraphics[width=.37\linewidth]{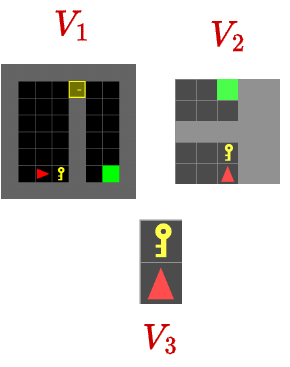}
		\caption{}
		\label{fig:views}
	\end{subfigure}%
	\begin{subfigure}{.65\textwidth}
		\centering \includegraphics[width=.65\linewidth]{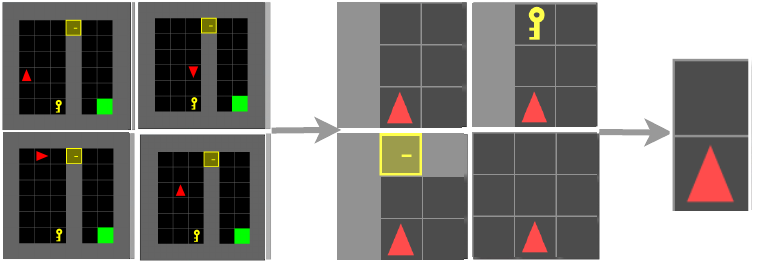}
		\caption{}
		\label{fig:mappings}
	\end{subfigure}
	\caption{\textbf{Hierarchical views allow us to transfer relevant invariances
      about the environment.} (a) The three different representations are
    obtained by cropping the observation for each state-action pair $(s, a)$.
    $V_1$ is the full view where the agent sees the whole environment or the
    largest portion of it. $V_2$ is an intermediate cropped representation of
    the agent's view that helps the agent generalize by reusing familiar
    observations . $V_3$ is the most restricted view where the agent only sees
    what is immediately in front of it. (b) Through the views $V_1$ to $V_k$ the
    amount of cycles continuously increases as the observations in the higher
    views can be mapped multiple times to the same observation in the lower
    view. This naturally leads to each view being a separate POMDP describing
    the true MDP.}
	\label{fig:views_mappings}
\end{figure*}

%
To discuss the roles of the different views more precisely, let's consider three
such cropped views which consist of the global view $V_1$, the intermediate view
$V_2$ and the smallest possible view $V_3$ (see Figure \ref{fig:views}). Each
view gives a new, typically more limited, perspective of the true state. As
shown in Figure \ref{fig:mappings}, each view induces a different set of cycles,
for instance a sequence of actions which do not lead to a cycle in the full view
$V_1$, might lead to a cycle in a smaller view $V_3$, because in the latter,
more states are mapped to the same observation. Thus, the different views
provide different types of information which are useful for learning and allow
the agent to focus on different properties of the environment.

In general, we can have an arbitrary number of views. Each view $V_i$ induces a
POMDP
\begin{align}
	\mathcal{V}_i &= (\mathcal{S}, \mathcal{A}, \mathcal{O}^{V_i}, \mathcal{P}, \mathcal{R}, \mathcal{Z}^{V_i}, \gamma)
\end{align}
each having their own set of observations $\mathcal{O}^{V_i}$ and observation
function $\mathcal{Z}^{V_i}: \mathcal{S} \rightarrow \mathcal{O}^{V_i}$. All
POMDP's $\mathcal{V}_1, \mathcal{V}_2, \ldots$ operate on the same state space
$\mathcal{S}$, however they have different sets of observations $\mathcal{O}$
and corresponding observation functions. General POMDPs can have a probabilistic
observation function. In our case, the observations are deterministic functions
of the full view, e.g. the observations for the $i$th POMDP of state $s$ is
\begin{align}
	o^{V_i} = \mathcal{Z}^{V_i}(s)
\end{align}
which corresponds to the cropping for the view $V_i$. Hereby we create partial
representations of the state space that allow us to identify invariances in the
environment by looking at the same true state $s$ through different
perspectives.

Note that, POMDPs are normally used to model uncertain observations. That is,
they are a generalization of MDPs where the true state is not observable. Here,
the POMDP idea is used to model different views of a fully observable state
space. Therefore, we do not seek to solve a POMDP problem, but rather we extend
a regular MDP to get redundant hierarchical representations.

\subsection{A cyclophobic policy for hierarchical state representations}
\label{sec:cycl-policy-hier}
In the following, we describe how the cyclophobic intrinsic reward and the
hierarchical state representations can be combined into a policy that exploits
the cyclophobic inductive bias. For this, we define several Q-functions along
the views, and combine them as a weighted sum. The weights are determined by
counts of the observations in each observation set $\mathcal{O}^{V_i}$ which we
explain next.
	
\paragraph{Mixing coefficients.}
Many strategies for defining mixing coefficients are possible. For concreteness,
in this paper we follow a simple schema, where we determine the weights from the
observation counts, which are obtained from the history of states visited
throughout training,
\begin{align}
  \mathcal{H}_\text{all} = \{ s_1, s_2, \ldots, s_T\}.
\end{align}
Note that $\mathcal{H}_\text{all}$ contains all states that have been visited in
the training so far, which is different from the states in the episodic history
$\mathcal{H}_\text{episodic}$ that was used in Sec.~\ref{sec:cir} to define
cycles. Denoting the corresponding views of the history as
$o_t^{V_i}=\mathcal{Z}^{V_i}(s_t)$, the counts for view $V_i$ are
\begin{align}
  N(o_1^{V_i}), N(o_2^{V_i}), \ldots, N(o_T^{V_i}).
\end{align}
where $N$ counts the number of times the observation $o_t^{V_i}$ has been
encountered. The raw counts are normalized by their maximum (maximum for
simplicity, other normalizations are possible), because the smaller views have
higher counts than the bigger views. The weights should be large for states that
have been seen less often, so we subtract the normalized counts from one,
\begin{align}
  \alpha_*^{V_i}(o_t^{V_i}) = 1 - \frac{N(o_t^{V_i})}{\max(N(o_1^{V_i}), N(o_2^{V_i}), \ldots, N(o_T^{V_i}))}.
\end{align}
The previous formula allows us to compute the weights for each respective view
$V_i$ for a single state $s_t$,
\begin{align}
  \alpha_*(s_t) = [\alpha_*^{V_1}(\mathcal{Z}^{V_1}(s_t)), \ldots, \alpha_*^{V_k}(\mathcal{Z}^{V_k}(s_t))] = [\alpha_*^{V_1}(o_t^{V_1}), \ldots, \alpha_*^{V_k}(o_t^{V_k})]
\end{align}
where $\alpha_*(s_t)$ is a vector. Finally, the softmax operator turns these
weights into a vector of probabilities,
\begin{align}
      \alpha(s_t) &= \text{softmax}(\alpha_*(s_t)) 
                    = [\alpha_1(s_t), \ldots, \alpha_k(s_t)].
    \end{align}
The entries of this vector are denoted by $\alpha_i(s_t)$ and are used
to define the cyclophobic Q-function in the next section.  The
definition of $\alpha$ can be extended to all state $s$ by setting it
to zero for unseen states, i.e. $\alpha(s)=0$ (zero-vector) for $s\not\in\mathcal{H}_\text{all}$.
    
In general, larger views in the hierarchy will have bigger entries in
$\alpha(s_t)$ as the observations repeat less often than in the
smaller views. Thus $\alpha(s_t)$ gives more weight to the larger
views than the smaller ones.  However, the converse is true for the
cyclophobic intrinsic reward, since in the smaller views the cyclophobic penalty is given far more often than in the larger views and thus
the q-values in smaller views will be higher than in the larger views.

\paragraph{Cyclophobic Q-function.}
To combine all views into a single global policy we define a mixture
over the different Q-functions learned with the cyclophobic intrinsic
reward as defined in Section \ref{sec:cir}. For view $V_i$, we define
Q-function as
\begin{align}
  Q(o^{V_i}, a) \leftarrow (1-\eta) \ Q(o^{V_i}, a) + \eta \big[r(o^{V_i},a) + \gamma Q(o'^{V_i}, a')\big].
\end{align}
This follows from our argumentation in Section \ref{sec:hsr}, where we
replace the state $s$ in Equation \ref{eq:qfunction} by the
observations $o^{V_i}$ in their respective views. Then we can define
cyclophobic Q-function as the mixture of the Q-functions of each view,
\begin{align}\label{eq:3}
  Q_{\text{cycle}} \big(s, a \big) = \sum_i  \alpha_i(s) \, Q(\mathcal{Z}^{V_i}(s), a) = \sum_i  \alpha_i(s) \, Q(o^{V_i}, a).
\end{align}
Note that the mixing coefficients $\alpha_i(s_t)$ are only
non-zero for states $s_t$ that appeared in the global history
$\mathcal{H}_\text{all}$.  Thus the cyclophobic Q-function is zero for
states $s\not\in\mathcal{H}_\text{all}$ not encountered.

\paragraph{Cyclophobic policy.} Finally, the cyclophobic policy is defined as
the greedy action for the cyclophobic Q-function, i.e.
\begin{align}
  \pi(s) = \argmax_{a} Q_{\text{cycle}} \big(s, a \big) = \argmax_{a} \sum_i \alpha_i(s) Q(o^{V_i}, a).
\end{align}
Having normalized the counts within each view ensures comparability of the
counts. This ensures that Q-values from rare i.e more salient observations have
a larger effect on deciding the action for the policy $\pi$. In an ablation
study in Section \ref{sec:experiments} we show that the combination of the
cyclophobic intrinsic reward and hierarchical state representations is crucial
to the method's success.

\section{Experiments}
\label{sec:experiments}
Our experiments are inspired by \cite{ioca} and
\cite{minihack_baselines}. We test in environments where the causal
structure is complex and the binding problem \citep{binding}, \citep{objects} arises. That is, where
some form of disentangled representation of the environments plays an
important role for efficiently finding solutions. 
\paragraph{Environments:} We test our method on the MiniGrid and MiniHack
environments:
\begin{itemize}
\item The MiniGrid environment \citep{gym_minigrid} consists of a series of
  procedurally generated environments where the agent has to interact with
  several objects to reach a specific goal. The MiniGrid environments pose
  currently a benchmark for the sparse reward problem, since a reward is only
  given when reaching the final goal state.
\item The MiniHack environment \citep{samvelyan2021minihack} is a graphical
  version of the NetHack environment \citep{nethack}. We select environments
  from the \emph{Navigation} and \emph{Skill} tasks. The MiniHack environment
  has a richer observation space by containing more symbols than the MiniGrid
  environments and a richer action space with up to 75 different actions. While
  not necessarily tailored to the sparse reward problem as the MiniGrid
  environment, the high state-action complexity makes it one of the most
  difficult environments for exploration.
\end{itemize}

\paragraph{State encoding:} For both environments we choose five croppings of
the original full view. The views $V_1, V_2, V_3, V_4, V_5$ are of grid size
$9 \times 9$, $7 \times 7$, $5 \times 5$, $3 \times 3$ and $2 \times 1$. In
principle we could also include the full view. However, in the experiments the
performance was much better when we limit ourselves to the partial views.
Furthermore, every part of the grid that is behind of the wall from the agent's
perspective is occluded (note that this is not the case in Figure
\ref{fig:views} for visualization purposes). Intuitively, limiting the views
allows the agent to ignore irrelevant details that are far away.

Next, the views are mapped to hashcodes (we use the open source \emph{xxhash}
library). That is, we have a hash function $g$ that maps observation $o_t^{V_i}$
to a hashcode $h_t^{V_i} = g(o_t^{V_i})$. This helps us to quickly check for
cycles as we only need to check whether two hashcodes are equal. For the
MiniHack environment, a text prompt is an integral part of the current state.
So, for MiniHack, the hashcode for a state is the concatenation of the hashcodes
of the observation $o_t^{V_i}$ and the text prompt $m_t$, where
$m_t \in \mathbb{R}^k$ is an encoding of a string of length $k$, i.e.
\begin{align}
  h_t &= g(o_t) + g(m_t). & \text{("$+$" denoting concatenation)}
\end{align}
Overall, while the observations in the MiniGrid and MiniHack environments are
more simple than in for instance Atari \citep{ale} or Habitat \cite{habitat},
the action-space and task that have to be solved are more complex in the
MiniGrid and MiniHack environments.

\paragraph{Training setup and baselines:} We train each agent for three runs
using different seeds for every run. For transfer learning, we use the
``DoorKey-8x8'' and MultiEnv(``MultiRoom-N4-S5'', ``KeyCorridorS3R3'',
``BlockedUnlockPickup'') setups for pretraining. For each environment we
pretrain for 5 million steps. For the MultiEnv setup this means that we pretrain
for a total of 15 millions steps which is less than \cite{ioca}'s 40 million
steps. For transfer learning, during pretraining we save the extrinsic rewards
in a second separate Q-table in addition to the main Q-table which contains
values of equation \ref{eq:qfunction}. These extrinsic rewards in the second
Q-table are then used at the beginning of transfer learning for each
environment, while we continue to use the cyclophobic intrinsic reward when
doing transfer learning. The baselines for the MiniGrid experiments are provided
by \cite{ioca} and allow us to compare our method to C-BET \citep{ioca}, Random
Network Distillation \citep{pseudo}, RIDE \citep{ride} and Curiosity-driven exploration \citep{icm}.
Furthermore, we also compare to NovelD \cite{noveld}. For the MiniHack
environment we compare our results to the baselines presented by
\cite{minihack_baselines} which include IMPALA \citep{impala}, RIDE \citep{ride}
and Random Network Distillation \citep{rnd}. For the skill tasks the only
available baseline is IMPALA \citep{impala}.

\paragraph{Evaluation metric:}
While during learning the intrinsic reward based on cyclophobia plays the
essential role, the ultimate goal is to maximize the extrinsic reward that is
provided by the environment. Thus for comparison, we have to plot the extrinsic
reward the agent receives for each episode. The reward ranges from 0 to
1. 

\subsection{MiniGrid}

\begin{figure*}[htb!]
  \centering \includegraphics[scale=0.32]{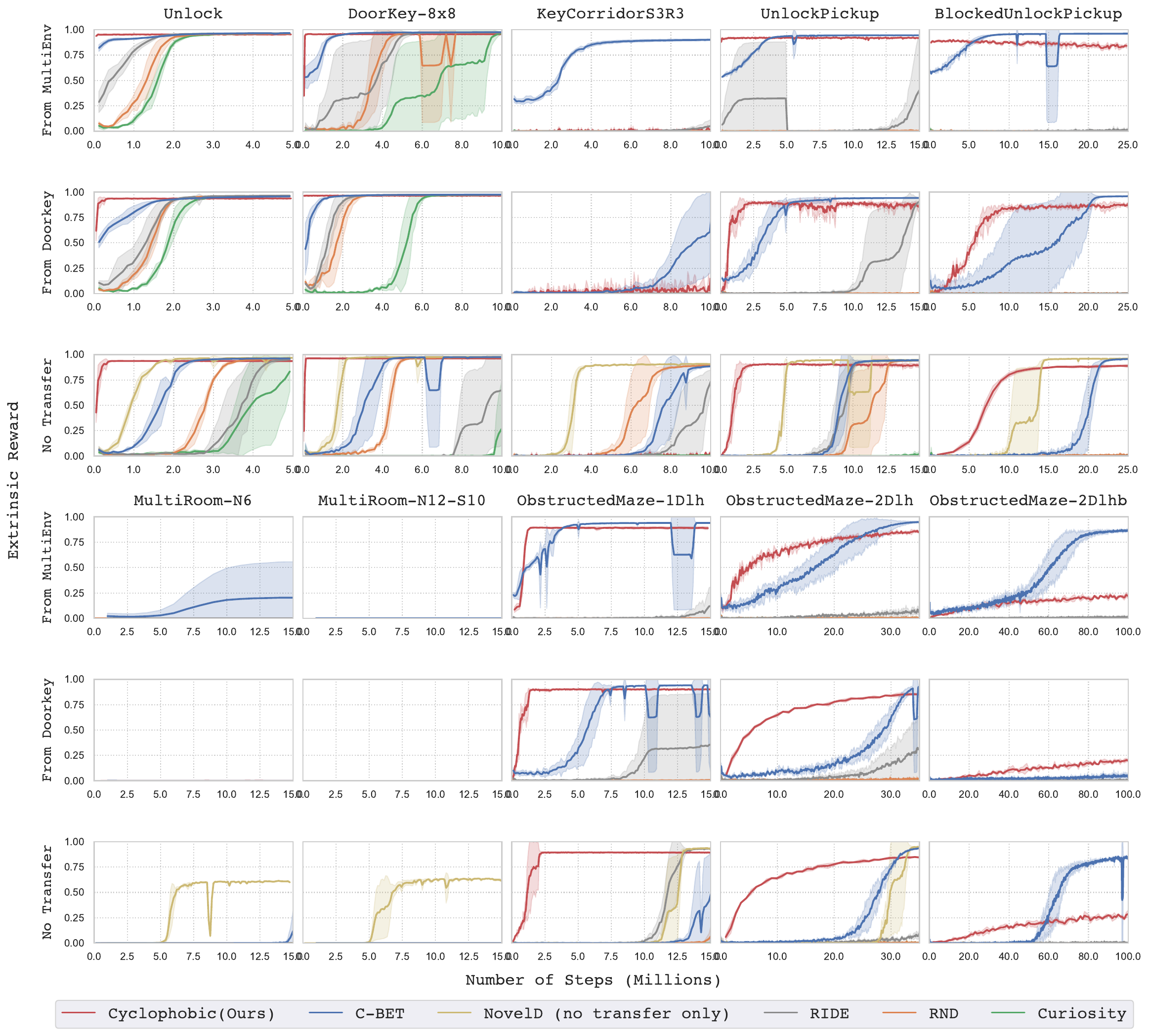}
  \caption{\textbf{MiniGrid:} We converge faster than C-BET \citep{ioca} in many
    MiniGrid environments with and w/o pretraining. We also converge faster than
    NovelD \cite{noveld} which we compare to in the ``no transfer'' setting
    only. The hierarchical state representations and cyclophobic intrinsic
    reward is extremely quick to converge which shows efficient exploration and
    the usefulness of the cropped representations, since we are also able to
    transfer knowledge by pretraining on other environments.}
  \label{fig:resultsmg1}
\end{figure*}
To test our method in the MiniGrid environment we choose the same training setup
of \cite{ioca}. That is, we determine the performance of the agent with no
pretraining and when pretraining from other environments as explained in the
previous section. Figure \ref{fig:resultsmg1} shows the agent's performance when
training from scratch (rows three and six) and when transferring knowledge from
pretrained environments (bottom three rows).
\begin{itemize}
\item \textbf{Learning from scratch} (rows three and six): In three out of six
  environments, our proposed method converges much faster than the competitors,
  including C-BET \citep{ioca} and NovelD \cite{noveld}. Only
  ``KeyCorridorS3R3'', ``MultiRoom'' and ``ObstructedMaze-2Dlhb'' pose
  significant challenges to our approach, because our method is tabular and thus
  cannot deal with too many object variations in the environment (e.g. random
  color changes). This is discussed in Section \ref{sec:limitations}.
  Furthermore, the ``MultiRoom'' environment proves challenging for all
  environments with only C-BET and NovelD managing to reach convergence, while
  we are able to fetch some rewards. This is due to the large amount of
  observations the different corridors produce. Note that we were not able to
  reach convergence in ``ObstructedMaze-2Dlhb'' with NovelD. In Figure
  \ref{fig:ablation_color2} our approach also excels in the ``KeyCorridorS3R3''
  and more difficult environments once we remove the colors, e.g.
  ``KeyCorridorS4R3'', ``KeyCorridorS5R3'', ``KeyCorridorS6R3''.
\item \textbf{Transferring knowledge} (rows one, two, four and five): Having
  trained on one environment can we transfer knowledge to a different one? In
  some environments the transfer even improved the results from the ``no
  transfer'' setup (see ``Unlock'', ``Doorkey'', ``UnlockPickup'',
  ``BlockedUnlockPickup'', ``ObstructedMaze-1Dlh'') and never deteriorated
  performance.
\end{itemize}

\subsection{MiniHack}
\begin{figure*}[htb!]
  \centering
  \includegraphics[scale=0.36]{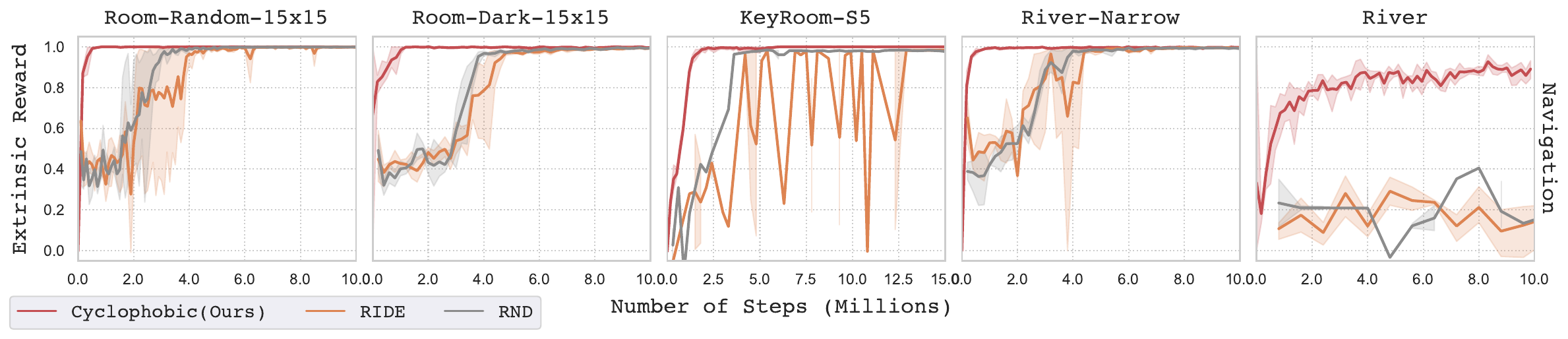}
  \caption{\textbf{MiniHack Navigation:} The agent converges quicker in the
    \emph{Navigation} task than the intrinsic curiosity baselines such as RIDE
    \citep{ride} and Random Network Distillation \citep{rnd}. This corroborates
    our hypothesis, that avoiding cycles is essential for quick exploration.}
  \label{fig:mh_results1}
\end{figure*}
\vspace{-1mm}
\begin{figure*}[htb!]
  \centering \includegraphics[scale=0.36]{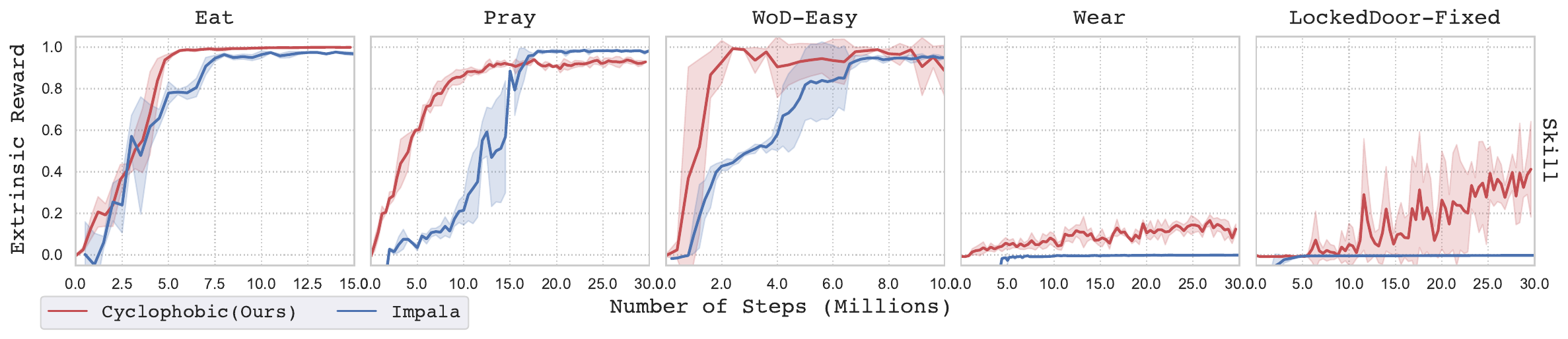}
  \caption{\textbf{MiniHack Skill:} We converge quicker than IMPALA
    \citep{impala} in the \emph{Skill} task. The \emph{Skill} task defines over
    75 different actions the agent must learn to use making it one of the most
    difficult sets of environments of the MiniHack suite.}
  \label{fig:mh_results2}
\end{figure*}
To push our approach to its limits, we also tackle some of the MiniHack
environments which requires the agent to learn a large skill set.
\begin{itemize}
\item \textbf{Navigation task:} For the simpler navigation tasks shown in Figure
  \ref{fig:mh_results1}, our method converges quicker than intrinsic curiosity
  driven methods such as RIDE and RND. Especially, in the ``River'' environment,
  only our cyclophobic agent is able to converge and solve the environment.
  However, of course there are environments such as ``RoomUltimate'' where our
  approach fails due to its tabular style, which limits the complexity of the
  environment.
\item \textbf{Skill task:} For the \emph{Skill} task the only available baseline
  is IMPALA which is not based on intrinsic curiosity. Here we are also vastly
  superior, even collecting extrinsic rewards when the baseline can not
  (``Wear'' and ``LockedDoor-Fixed'').
\end{itemize}

\subsection{Ablation Study}
\begin{figure}[htb!]
  \centering \hspace*{-0.2cm} \includegraphics[scale=0.29]{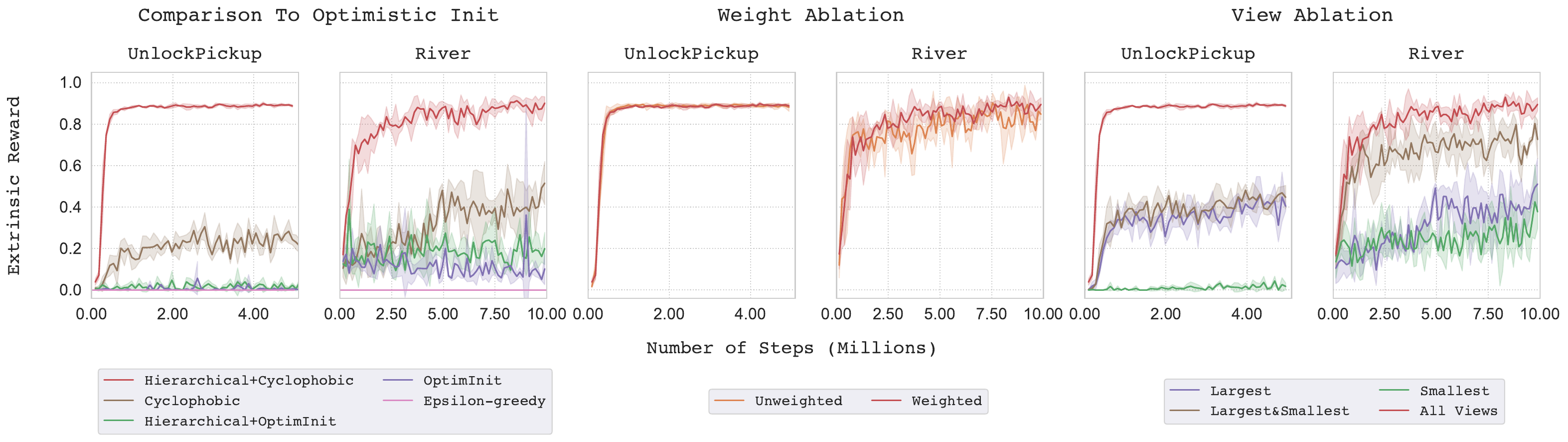}
  \caption{\textbf{Hierarchical state representations ablation:} Hierarchical
    state representations are crucial to performance in cyclophobic
    reinforcement learning. In the first two figures we compare cyclophobic
    reinforcement learnign to optimistic initialization with and without
    hierarchal views and see that hierarchical views improve performance for
    both the optimisitic initialization and cyclophobic agent. In the two middle
    figures we measure the effect of weighting the different views. In the last two figures we toggle between the largest, intermediate
    and smallest views. We see that all 3 types of views deliver the best
    performance.} 
  \label{fig:ablation_h}
\end{figure}

\begin{figure*}[htb!]
	\centering \hspace*{-0.2cm}
  \includegraphics[scale=0.28]{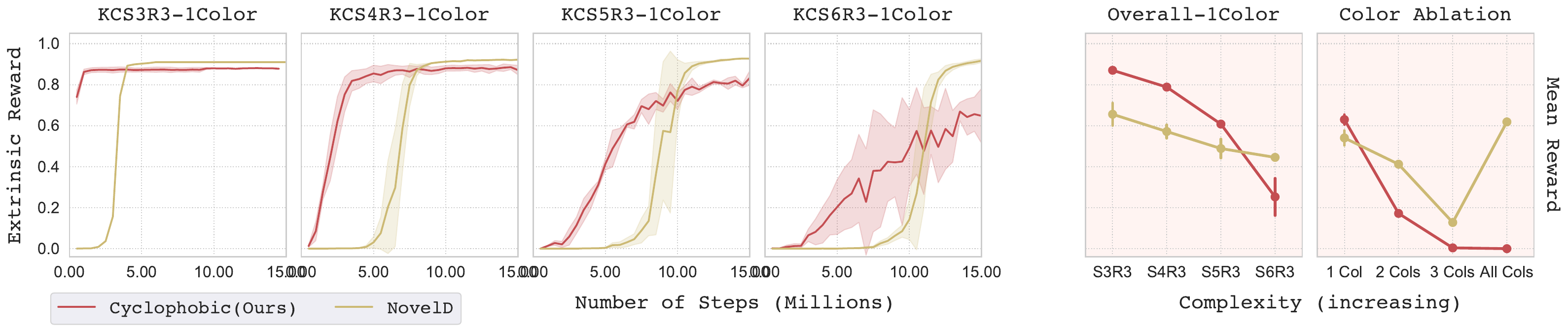}
	\caption{\textbf{Observational complexity ablation study (reducing colors):}
    The cyclophobic agent can gather rewards successfully in large
    ``KeyCorridor'' environments when reducing colors, showing that the
    exploration mechanism is very strong. Generally, the more distinct i.e. complex the observations in the environment are, the more the learned invariances from the neural network matter.}
	\label{fig:ablation_color2}
\end{figure*}

\begin{figure*}[htb!]
  \centering
  \hspace*{-1.7cm}
  \includegraphics[scale=0.15]{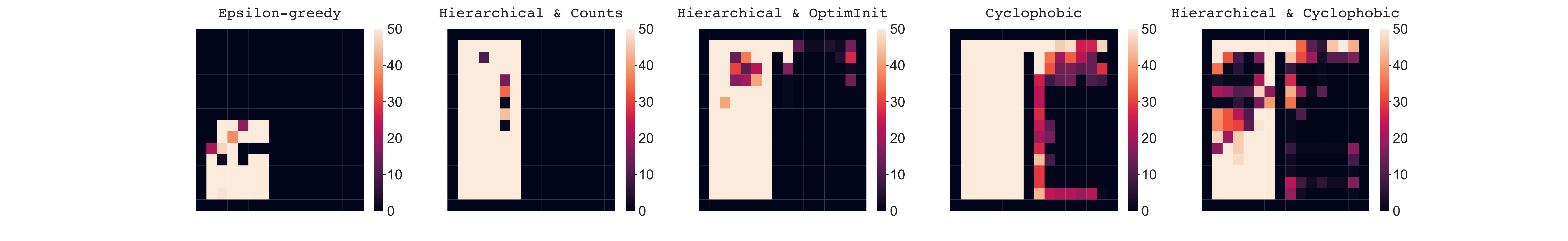}
  \vspace{-5mm}
  \caption{\textbf{Exploration behaviour(visitation counts for
      ``DoorKey-16x16''):} The cyclophobic Q-function explores the environment
    more efficiently than the optmistic initialization, count and epsilon-greedy
    based counterparts. We record visitation counts for several variations of
    intrinsic reward and hierarchical state representations. Hierarchical state
    representations together with the cyclophobic intrinsic reward are the most
    efficient. }
  \label{fig:ablation_heatmap}
\end{figure*}

We analyze several aspects of cyclophobic reinforcement learning in our ablation study. We first take a look at the impact of hierarchical state representations on the performance of cyclophobic reinforcement learning (see Figure \ref{fig:ablation_h}). Furthermore, we study how observational complexity affects exploration by ablating a reward-independent observational feature in the observations, that is we reduce the number of different colors for objects in the ``KeyCorridor'' environment (see Figure \ref{fig:ablation_color2}). Finally, we also analyze how efficiently the cyclophobic intrinsic reward explores the environment by visualizing state visitation counts (see Figure \ref{fig:ablation_heatmap}).

\paragraph{Impact of hierarchical views:} In Figure \ref{fig:ablation_h} we
analyze the impact of the hierarchical state representations on the agent's
performance. The training setup is described in section \ref{sec:pls}. We ablate
the hierarchical views in 3 ways. (i) In the first two figures we measure
performance of cylophobic reinforcement learning and optimistic initialization
with and without the hierarchical views. Both for optimistic initialization and
for cyclophobic reinforcement learning the hierarchical views improve
performance.
(ii) In the middle two figures we measure the effect of weighting
the different views according to their visitation counts. We see that for
``River'' the performance is improved while for ``UnlockPickup'' both weighted
and unweighted variants perform the same. Overall the effect of weighting the
different views is not very pronounced. (iii) Finally in the last two figures we
toggle the largest, intermediate and smallest views. Only using single views
performs the worse. Combining the largest and smallest views improves
performance, especially for ``River''. However, the best performance is achieved
when combining largest, intermediate and smallest views together.

\FloatBarrier
\FloatBarrier

\paragraph{Importance of state representations:} We demonstrated the impact of
learning or imposing a state representation for an agent with hierarchical state
representations. We now analyze how much of the success in exploration is due to
the function approximator learning observational invariances as described in
Section \ref{sec:limitations}. In Figure \ref{fig:ablation_color2} (white), we
fix the color of objects in the environment to only one color and compare our
method to NovelD \citep{noveld}, since it is able to solve all MiniGrid
environments. We see that in ``KeyCorridorS4R3'' and ``KeyCorridorS5R3'' our
method is more sample efficient than NovelD. Only in the ``KeyCorridorS6R3'' the
large state space reduces performance (see Figure \ref{fig:ablation_color2}
(pink left)). In general, we see that when the environment is observationally
less complex (e.g by removing colors (see Figure \ref{fig:ablation_color2} (pink
right)), the cyclophobic agent is more sample efficient than NovelD. However, as
soon as observational complexity increases NovelD takes the upper hand.

\FloatBarrier
\FloatBarrier

Overall, we observe that exploration success is not necessarily due to the intrinsic
reward itself, but can also largely be attributed to the learned
representations. Thus, when complexity in the observations increases, better
performance in neural networks can be attributed to the learned invariances
reducing observational complexity and not necessarily to a more efficient
exploration strategy. This can be seen in the tabular cyclophobic agent as well,
since the hierarchical state representations are crucial to solving the
environments.


\paragraph{Exploration behavior:}
\FloatBarrier
\FloatBarrier
To show the effectiveness of the cyclophobic intrinsic reward and the
hierarchical state representations, we perform an ablation study. Figure
\ref{fig:ablation_heatmap} shows state counts as a heat map after 10,000 steps
of training. We distinguish four cases: \textbf{(i) epsilon-greedy:} Plain
epsilon greedy exploration fails to find the goal in the bottom right.
\textbf{(ii) hierarchical \& counts:} We replace the cyclophobic intrinsic
reward in equation (3) with a count-based intrinsic reward similar to
\cite{mbie} defined by 
$N(o_T^{V_i})^{-\frac{1}{2}}$, for view $V_i$. This improves the results but
still fails. \textbf{(iii) hierarchical \& optimistic initialization:} We use
optimistic initialization to let the agent try all possible actions. We
initialize the q-table with a value of two. Optimistic initialization manages to
enter the second room but just fails at getting a reward within 10,000 steps.
\textbf{(iv) cyclophobic:} Having a cyclophobic intrinsic reward calculated only
on the largest view finds the goal. \textbf{(v) cyclophobic \& hierarchical:}
The combination of hierarchical views and cyclophobic intrinsic rewards (as
explained in Sec.~\ref{sec:cycl-policy-hier}) explores even more efficiently as
can be seen in the left room where fewer steps are needed to leave it.

\section{Related Work}
\label{sec:rel}

\paragraph{Breadth-first-search vs depth-first-search exploration:}
The way different exploration methods explore the state-action space is an
important aspect of designing exploration methods. We characterize this by
determining how an exploration method explores on the MDP-level. That is, the
way the next action is chosen and the visited states-action pairs that result
from the instrinsic reward. With count-based exploration \citep{pseudo,
  pseudo_ovr} where instrinsic reward are usually inverse visitation counts
$N(o_T^{V_i})^{-\frac{1}{2}}$, the more a state is visited the less reward it
will receive. In the short term, the agent will choose the same state-action
pair on its next visit (since it has been given an intrinsic reward for being
seen), until a random state-action pair is chosen. While this encourages the
agent to revisit previously seen states, this also leads to a depth first
approach to exploring the MDP and other action are only considered after
randomly being chosen by $\epsilon$-greedy. Another way to explore the
environment is by exploiting prioproceptive information in the representations.
Methods that use change based counts \citep{ioca, noveld} or prediction error
\citep{ride, icm, rnd, e3b} as intrinsic reward give the highest intrinsic
rewards to observations that are novel and have not been seen before. In this
case, the resulting exploration heuristic is based on pursuing the most novel
state-action pairs. This may also be problematic when the stochasticity in the
observations is not necessarily aligned with the task at hand. Overall, this
also equates to a depth-first-search on the MDP-level, since the agent will be
enticed to pursue the most novel transition instead of exploring other local
options.

In contrast, optimistic initialization \citep{dmoptim, dora} systematically
forces exploration on state-action pairs that have not been tried yet by the
agent. The idea of optimistic initialization is to initialize the value function
by a value larger than the expected reward. As the agent performs updates on the
value function it will be forced to choose the next unexplored action since the
value-function update for other state-action pairs will have decreased their
value. Methods like Never Give Up (NGU) \citep{ngu} try to combine both
depth-first-search and bread-first-search. NGU is composed of two networks where
one network will look for novel state transitions and another network tries to
give an intrinsic reward when a similar state is visited again to encourage
revisiting of states by doing a nearest-neighbor search. While NGU's focus is
more based on mitigating the lack of breadth-first-exploration in prediction
error based methods, cyclophobic reinforcement learning can be seen as an
extension of optmistic initialization where we incorporate more
depth-first-search into optimistic initialization with the hierarchical state
representations. Other methods such as RAM \citep{ram}, also attempt this by
defining an instrinsic reward based on the saliency of actions in different
states.

\paragraph{Loop closure, generalization and transfer learning:}
\cite{loopclosure} define loop closure as an auxiliary loss for exploration in
3D mazes in addition to a depth measure. The authors detect loop closure by
checking whether the agent has been near a previously visited position. To get
position information the authors integrate velocity information over time.
Furthermore, NovelD \citep{noveld} can also be seen as indirectly detecting
loops as it part of its intrinsic reward is based on only rewarding states when
they have not been visited before. Learning representations of the environment
that are reusable and can be transferred has been explored previously. Learning
successor representations \citep{successor} requires learning local environment
dynamics that can be reused when the environment, i.e. the original
distribution, is changed. \citep{uvfa} use Singular Value Decomposition on the
learned action-value function to obtain a canonical representation of the
environment which can be transferred to other situations. More recently,
\cite{ioca} learn an exploratory policy which is then combined with a task
specific policy at transfer time. Our method likewise contains a task-agnostic
component given by the hierarchical state representations and a task-specific
component given by the cyclophobic intrinsic reward.
		
\section{Discussion}

\subsection{Limitations}
\label{sec:limitations}
In some environments with a big number of different objects our method struggles
to converge. The tabular agent does not have the observational invariances that
methods based on function approximation learn and exploit (e.g.~with CNNs). For
instance, for these environments learned invariances allow them to handle
differently colored objects. While the tabular agent is able to avoid redundancy
through the cyclophobic intrinsic reward and generalize this to other parts of
the environment through the hierarchical state representations, it does not
learn the kind of observational invariances neural networks do, which are
sometimes necessary. However, note that we can show that reducing the number of
colors for the ``KeyCorridor'' environment improves our performance dramatically
(see Figure \ref{fig:ablation_color2}). 
Also, preliminary results show a cyclophobic PPO agent (see Appendix
\ref{sec:ppo}) solving ``KeyCorridor'', further supporting our theory. We
explore this in our ablation study. An additional limitation that results from
our tabular approach is that the entire observation history
$\mathcal{H}_{\text{all}}$ has to be recorded. This however can be mitigated by
using an unweighted sum of the q-values for all views as we show in the ablation
study.

\subsection{Contribution to existing literature}
At a first glance, cyclophobic reinforcement learning can be seen as another
count-based method where equal states are counted as cycles. Then instead of
giving an intrinsic reward which is the inverse of visitation counts, the
intrinsic reward is a fixed penalty given everytime a cycle is encountered i.e.
a state is counted. As explained in Section \ref{sec:rel}, the cycle penalty
encourages an exploration behaviour that explores all actions within a state
first instead of pursuing the least visited, i.e., most novel states. This is
inherently different from other count-based approaches. In this sense, the
cyclophobic intrinsic reward behaves like optimistic initialization where the
focus of exploration is placed on the actions instead of the states. Moreover,
the hierarchical state representations contrast the breadth-first exploration by
learning multiple value functions which are expressive due to the different
contents in the hierarchical views. Only few other works such as NGU have
followed this approach of balancing breadth-first and depth-first exploration as
described in Section \ref{sec:rel}.

\subsection{Extension to high-dimensional state spaces}
So far we have tested cyclophobic reinforcement learning on the MiniGrid and
MiniHack environments. While the state-action space of these environments is
combinatorially complex to solve from an exploration standpoint, the
observations are simple compared to other environments such as Atari \citep{ale}
or Habitat \citep{habitat} where the exploration task may not be as complex, but
the observations are noisy and high-dimensional. Since cyclophobic reinforcement
learning is a count-based method, we propose to extend cyclophobic reinforcement
learning to high dimensional observations by using density estimation techniques
as in \cite{pseudo, pseudo_ovr}. For the hierarchical state representations a
PixelCNN \citep{pixelcnn} or VQ-VAE \citep{vqvae} could be used to estimate the
density of the observations. With smaller views we expect these densities to get
easier to estimate as they contain less observational detail. For the
cyclophobic penalty, a pseudo count could be introduced as in \cite{pseudo,
  pseudo_ovr}. Another approach would be to define an embedding network as in
NGU \citep{ngu} and searching for nearest neighbors in the episodic history to
detect cycles. Note that in NGU's case the embedding network is used to revisit
states while in our case it would be used to detect cycles in order to
discourage visiting these states. Overall, we leave the extension to
high-dimensional state spaces for future work.

\subsection{Conclusion}
Avoiding cycles allows the agent to quickly and systematically discard
uninteresting state-action pairs which are repeated often. This makes our
cyclophobic intrinsic reward a good inductive bias towards novelty that
simultaneously encourages systematic exploration. The ablation studies show that
the cyclophobic intrinsic reward just for a single view is already powerful
enough to solve complex environments. Adding hierarchical state representations
leads to even better performance, as can be seen in the MiniGrid experiments.
Moreover, the experiments in MiniGrid confirm the transferability of the learned
bias to new environments. Here, the hierarchical state representations are
crucial since we detect cycles in smaller views that can be generalized to
larger ones. In addition, the policy is task-agnostic since the cropped views
can be easily transferred to other environments.
Due to the tabular approach, environments with higher observational complexity
lead to convergence problems with our method.
       Neural networks, learn invariances that reduce the complexity of the
       state space. To address the convergence problems, future work will
       therefore focus on incorporating cyclophobic reinforcement learning into
       a neural network based
       architecture. 

       \bibliography{references}

       \bibliographystyle{tmlr}

       \newpage
\appendix
\section{Further Ablation Studies}
For all plots in the ablation studies we use the same setup as in the main experiments. We refer to section \ref{sec:pls} for this.
\subsection{Preliminary results with PPO}
\label{sec:ppo}
\FloatBarrier
\begin{figure}[htb!]
	\centering \includegraphics[scale=0.44]{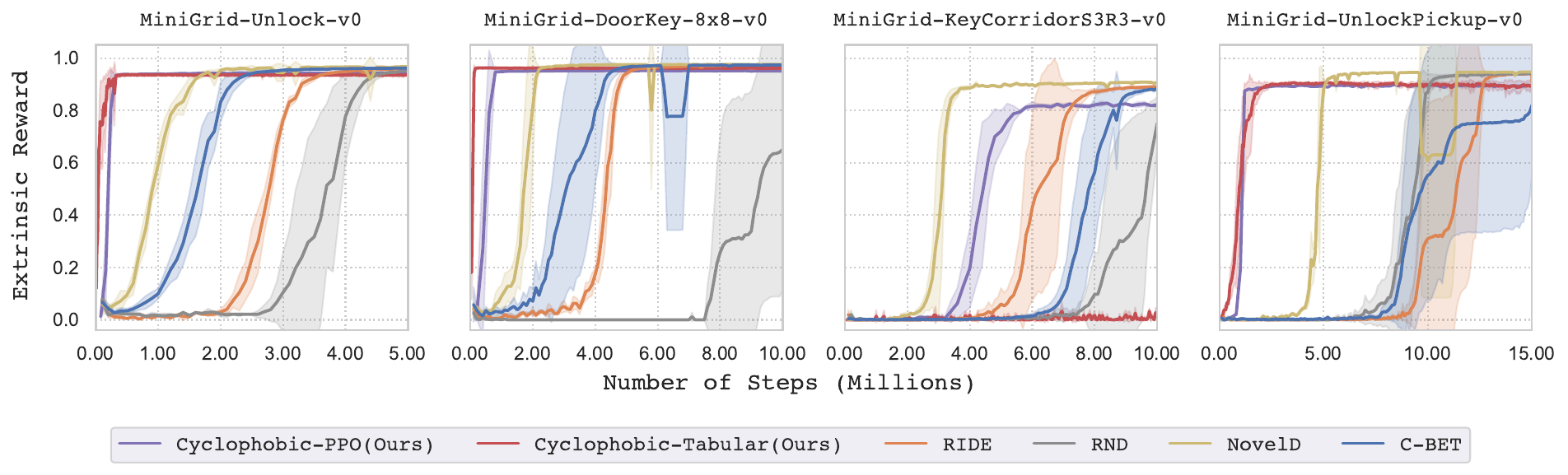}
	\caption{Using PPO together with only the cyclophobic intrinsic reward also
    converges far quicker than C-BET. Moreover, we are able to converge in
    KeyCorridorS3R3 confirming our theory in Section \ref{sec:limitations}.}
	\label{fig:nn}
\end{figure}
\FloatBarrier We train a PPO agent using only the cyclophobic intrinsic reward.
Figure \ref{fig:nn} shows that for the given environments the PPO agent also
explores more efficiently than C-BET. Furthermore, our theory about observation
complexity from \ref{sec:limitations} is supported as the PPO agent is able to
solve the ``KeyCorridor'' environment. Note that these are preliminary results.
In future work we seek to implement the hierarchical state representations for
the neural agent.

\section{Environment Details}
\subsection{MiniGrid Experiments}\label{sec:minigrid-experiments}
\FloatBarrier
\begin{figure}[htb!]
	\centering
	\includegraphics[scale=0.52]{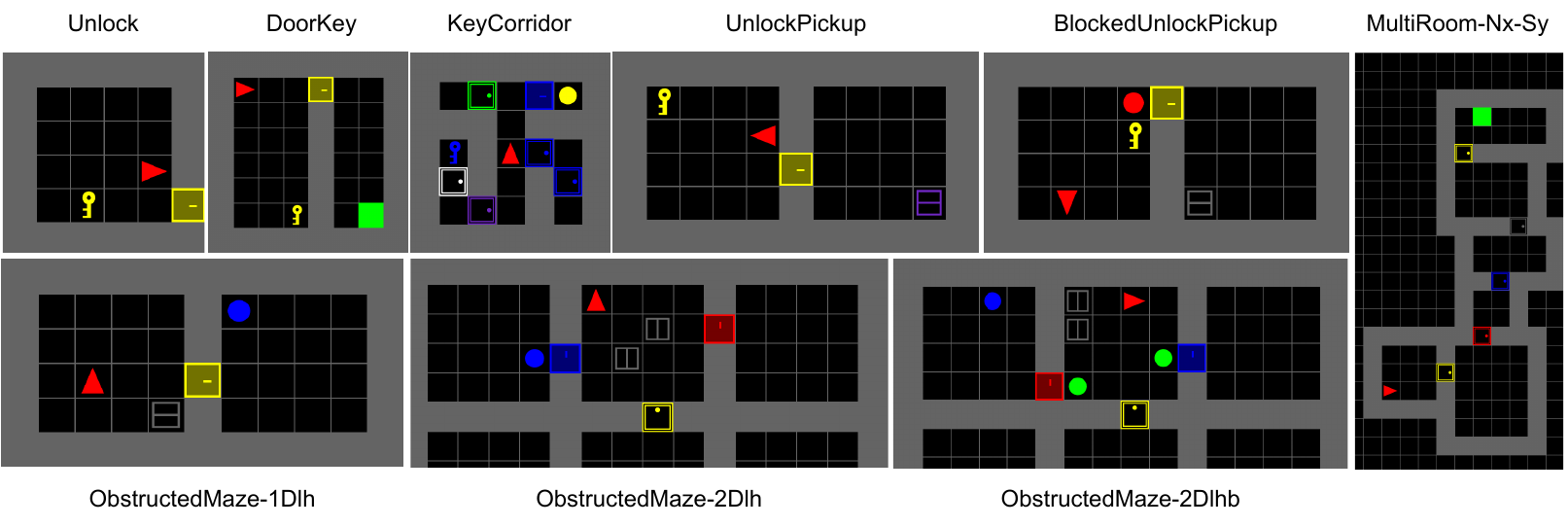}
	\caption{An overview of the MiniGrid environments used in this paper.}
	\label{fig:minigrid_overview}
\end{figure}
\FloatBarrier
We test on some of the same environments as \cite{ioca}. In the following we describe the environments as \cite{ioca} and add our own comments where relevant.
\begin{itemize}
	\item \textbf{Unlock}: pick up the key and unlock the door. (288 steps per episode)
	\item \textbf{DoorKey-8x8: }pick up the key, unlock the door, and go to green goal (640 steps per episode)
	\item \textbf{KeyCorridorS3R3:} pick up the key, unlock the door, and pick up the ball (only the door before the ball is locked) (270 steps per episode).
	\item  \textbf{UnlockPickup: }pick up the key, unlock the door, and open the box (288 steps per episode).
	\item \textbf{BlockedUnlockPickup:} pick up the ball in front of the door, drop it somewhere else, pick up the key, unlock the door, and open the box (576 steps per episode).
	\item \textbf{ObstructedMaze-1Dlh:} open the box to reveal the key, pick it up, unlock the door, and pick up the ball (288 steps per episode). 
	\item \textbf{ObstructedMaze-2Dlh:} same as above, but with two doors to unlock (576 steps per episode). 
	\item \textbf{ObstructedMaze-2Dlhb: }same as above, but with two balls in front of the doors (like in BlockedUnblockPickup) (576 steps per episode).
	\item \textbf{ MultiRoom-N6:} navigate through six rooms of maximum size ten and go to the green goal (all doors are already unlocked) (120 steps per episode). 
	\item \textbf{ MultiRoom-N12-S10:} navigate through twelve rooms of maximum size ten and go to the green goal (all doors are already unlocked) (120 steps per episode). 
\end{itemize}

\subsection{MiniHack Experiments}\label{sec:minihack-experiments}
\FloatBarrier
\begin{figure}[htb!]
	\centering
	\includegraphics[scale=0.46]{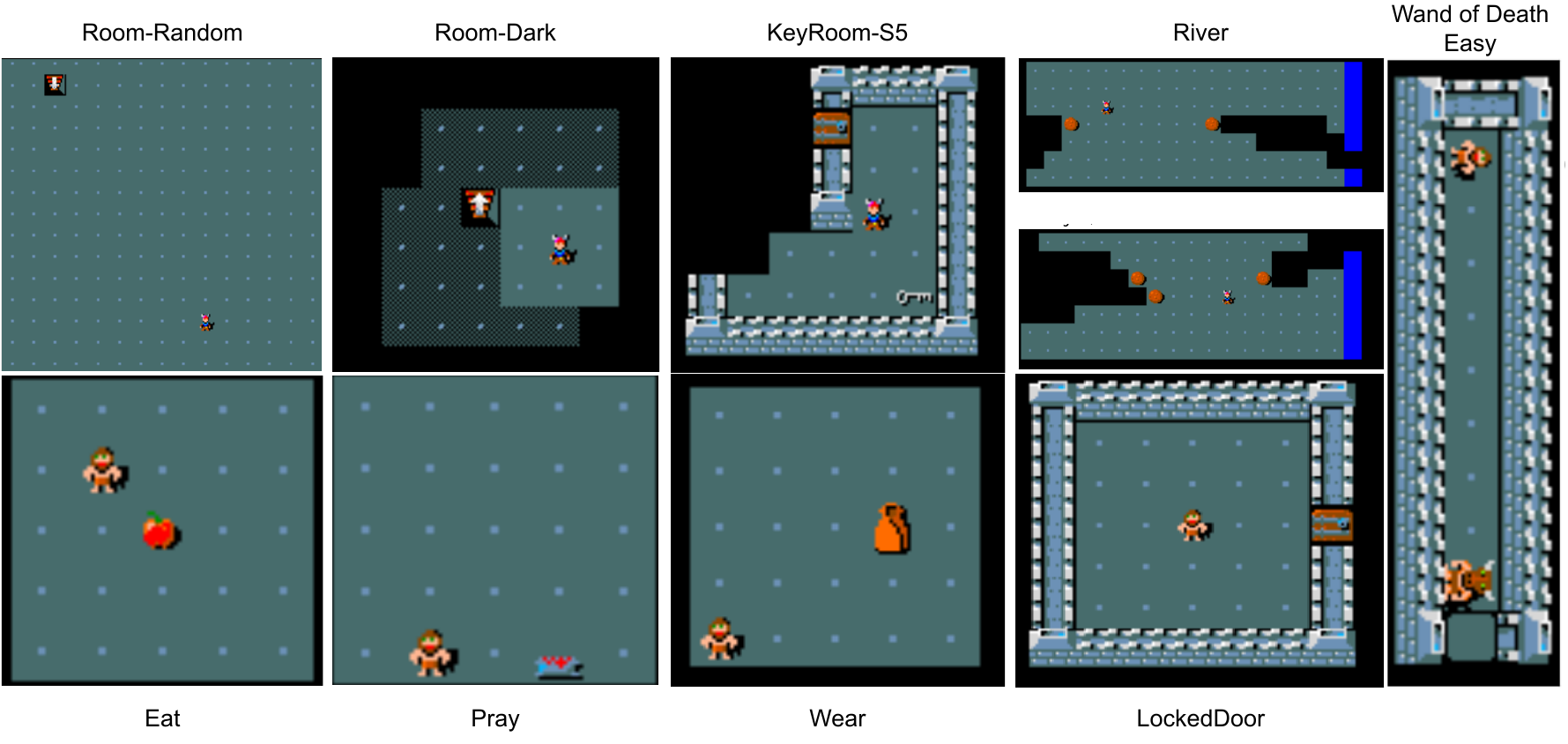}
	\caption{An overview of the Minihack environments tested in this paper. The top row shows the \emph{Navigation} tasks, while the bottom row shows the \emph{Skill} tasks. }
	\label{fig:minihack_overview}
\end{figure}
\FloatBarrier
We perform experiments inspired by the baselines from \cite{minihack_baselines}. Here, \cite{minihack_baselines} define a set of navigation and skill acquisition tasks.  The navigation tasks range from requiring the agent to navigate an empty room or maze, up to navigating environments with monsters and hazards. The skill acquisition tasks require the agent to perform a certain set of skills to be able to solve the environments.
\subsubsection{Navigation tasks}
In the following we describe the environments used from the navigation task.
\begin{itemize}
	\item \textbf{Room-Random-15x15:} This task is set in a single square room, where the goal is to reach the staircase down. In the random version of this task, the start and goal positions are randomized.
	\item\textbf{ Room-Dark-15x15:} Same as Room-Random, but the agent only sees parts of the environment it has already discovered.
	\item \textbf{KeyRoom-S5:} These tasks require an agent to pick up a key, navigate to a door, and use the key to unlock the door, reaching the staircase down within the locked room. The locations of the door, key and starting position are randomized.
	\item \textbf{River:} The river environment requires the agent to cross a river using boulders. When pushed into the water, they create a dry land to walk on, allowing the agent to cross it.
\end{itemize}
\subsubsection{Skill tasks}
The nature of commands in NetHack requires the agent to perform a sequence of actions so that the initial action, which is meant for interaction with an object, has an effect. The exact sequence of subsequent can be inferred by the in-game message bar prompts. Overall, in each of the following tasks the agent has to perform a sequence of actions that allow the agent to perform the required skill.
\begin{itemize}
	\item \textbf{Eat: }The agent has to navigate and eat an apple. Doing so requires a sequence of actions and confirming a prompt to eat the appe. 
	\item \textbf{Pray:} The agent has to pray on an altar. This is more complex than eating an apple as praying requires the agent to confirm more prompts.
	\item \textbf{Wear: }The agent has to wear armor. This is a more difficult task than the previous tasks. 
	\item \textbf{LockedDoor-Fixed:}The agent has to kick a locked door.
	\item \textbf{Wand of Death-Easy:} The agent starts with a wand of deatch in its inventory and has to zap it towards a sleeping monster. That is, the agent needs to equip the wand of death walk to the monster and zap it.
\end{itemize}
In all of these environments the positions of the agent and of the object that has to be interacted with are randomised. Furthermore, the complexity is also increased by the confirmation prompts the agent has to answer in order to perform a certain action. 

\section{Experimental Details}
\subsection{Hyperparameters}
\paragraph{Common hyperparameters:} In the following we present a table with all hyperparameters used for training the environments. In general, our method has very few hyperparameters. In this case the only really tunable hyperparameters are the step size $\eta$, the random action selection coefficient $\epsilon$ for epsilon-greedy exploration and the discount $\gamma$:
\begin{center}
	\begin{tabular}{ l | c | r }
		\hline
		$\eta$ & $\gamma$ & Num. views \\ \hline
		0.2 & 0.99 & 5 \\ 
		\hline
	\end{tabular}
\end{center}
In our case we determined experimentally that 5 views is a good number of views for training an agent.
\paragraph{Epsilon-greedy parameter $\epsilon$:} While learning with the cyclophobic intrinsic reward we still perform random action from time to time by having an epsilon-greedy action selection. That is, the agent samples a value $n$ from a uniform distribution and selects a random action if this $n < \epsilon$. These are the settings for $\epsilon$ for the respective environments:
\begin{center}
	\begin{tabular}{ l | c  }
		\hline
		Environments & $\epsilon$ \\ \hline
		Unlock & 0.1 \\ 
		DoorKey & 0.1 \\
		KeyCorridor & 0.1 \\
		UnlockPickup & 0.3 \\
		MultiRoom-N6 & 0.1 \\
		MultiRoom-N12-S10 & 0.1 \\
		BlockedUnlockPickup & 0.3 \\
		ObstructedMaze-1Dlh & 0.3 \\
		ObstructedMaze-2Dlh & 0.1 \\
		ObstructedMaze-2Dlhb & 0.3 \\
    All MiniHack & 0.3 \\
		\hline
	\end{tabular}
\end{center}
\paragraph{Intrinsic reward tradeoff $\rho$:} The intrinsic reward tradeoff
$\rho$ has the effect that the extrinsic reward gets propagated more
when the agent encounters it, which is necessary for more complex environments.
$\rho$ is a hyperparameter that can be defined for each environment for best performance.
\begin{center}
	\begin{tabular}{ l | c  }
		\hline
		Environments & $\rho$ \\ \hline
		Unlock & 1.0 \\ 
		DoorKey & 1.0 \\
		KeyCorridor & 1.0 \\
		UnlockPickup & 2.0 \\
		MultiRoom-N6 & 2.0 \\
		MultiRoom-N12-S10 & 2.0 \\
		BlockedUnlockPickup & 5.0 \\
		ObstructedMaze-1Dlh & 2.0 \\
		ObstructedMaze-2Dlh & 5.0 \\
		ObstructedMaze-2Dlhb & 5.0 \\
    River                & 5.0 \\
    WOD-Easy             & 5.0 \\
    Eat                  & 2.0 \\
    Pray                 & 2.0 \\
    Rest of MiniHack     & 2.0 \\
		\hline
	\end{tabular}
\end{center}

\subsection{PPO Hyperparameters}
\begin{center}
	\begin{tabular}{ l | c  }
		\hline
		\hline
		GAE-$lambda$ & 0.95 \\
		$\gamma$ & 0.99 \\
		Batch size& 256 \\
		Learning rate & 0.001 \\
		Entropy coefficient & 0.01 \\
		Value loss coefficient & 0.5 \\
		Max Grad Normt & 0.5 \\
		Clipping-$\epsilon$ & 0.2 \\
		RMSProp-$\epsilon$ & 1e-8 \\
		RMSProp-$\alpha$ & 0.99 \\
		\hline
	\end{tabular}
\end{center}
\subsection{Plot smoothing}\label{sec:pls}
For all 3 runs we show the mean reward and standard deviation smoothed over a
sliding window. For our and \cite{minihack_baselines}'s experiments we use a
sliding window of 50000 steps. For the experiments in \cite{ioca} we use a
sliding window of 96000 steps as per their experimental setup. The shaded areas
are the standard deviation of the mean for 3 runs.



\end{document}